\documentclass[pdflatex,sn-basic]{sn-jnl}% Basic Springer Nature author-year style

\usepackage{amsmath,amssymb}
\usepackage{graphicx}
\usepackage{booktabs}
\usepackage{csquotes}

\begin{document}

\title[Beyond Hooking Onto the World]{Beyond Hooking Onto the World: Referential Profiles and the Numerical Structure of LLM Grounding}

\author*[1]{\fnm{Joo Yull} \sur{Rhee}}
\email{rheejy@skku.edu}

\affil*[1]{\orgdiv{Department of Physics}, \orgname{Sungkyunkwan University}, \orgaddress{\city{Suwon}, \country{Republic of Korea}}}

\abstract{
The problem of grounding has returned in a new form with large language models (LLMs). Classical debates asked how symbolic systems could acquire meaning rather than merely manipulate formal tokens. Recent work reformulates this issue as a vector grounding problem: if LLMs compute over vectors, then the relevant question is how vector states and vector-mediated outputs can become connected to the world. I accept this shift, but argue that it remains incomplete in two respects. First, the notion of reference is often too thin. Reference is sometimes described as a word or representation ``hooking onto'' an object or property, but this metaphor risks treating reference as a fixed link between an isolated expression and an isolated worldly item. Even in the human case, however, reference is not publicly available as a private mental hook. We attribute reference to others through patterns of linguistic and practical coordination: their ability to use expressions, distinguish cases, accept corrections, explain relations, reidentify topics, and continue discourse. Reference is therefore better understood as profile-based, context-sensitive, discourse-level, affectively shaped, and norm-governed. Second, the vector grounding debate has not sufficiently explained how such language-mediated referential profiles become numerically usable inside a trained model. I argue that LLMs do not acquire reference through human perception, episodic memory, embodiment, or understanding. Rather, through numerical optimization, they relationally parameterize linguistic traces of human world-directed practice. In a finite vector system, referential profiles must be distributed, may be superposed with other profiles, and are recovered through context-sensitive computation. Weights, activations, attention-mediated hidden states, softmax-trained contrasts, and inner-product alignments are the mathematical sites at which inherited linguistic relations become stable and causally active. Mechanistic interpretability findings, including entity-like features, knowledge neurons, and emotion-related activation directions, provide indirect but important evidence for this view. They do not show that LLMs possess human reference. They support a more limited thesis: LLMs may possess derivative, language-mediated, profile-based, and numerically structured forms of reference.
}

\keywords{large language models; vector grounding problem; symbol grounding problem; reference; referential profiles; causal-informational relations; relational parameterization; distributed representation; superposition; mechanistic interpretability; philosophy of AI}

\maketitle

\section{Introduction}
Large language models (LLMs) create a peculiar philosophical pressure. On one hand, they are just software systems running on computers. They do not perceive the world, remember as human beings remember, act with intentions, or take responsibility for what they say. In this respect, a language model is no less a machine than a word processor or a scientific computation code. It can do only what computation allows: transform inputs into numerical states, propagate those states through learned operations, and produce outputs. On the other hand, unlike a word processor, an LLM produces fluent, context-sensitive, and apparently world-directed language. It answers questions about cities, people, events, emotions, mathematical objects, laws, fictional characters, and scientific entities. It can distinguish Paris, France from Paris, Texas. It can treat ``Golden Gate Bridge'' not merely as three adjacent words, but as a name connected with a particular bridge, a city, a visual profile, a history, and a set of human practices.

This double attitude structures much of the contemporary debate. The mechanistic description seems to forbid attributions such as meaning, understanding, or reference: how could a numerical system that only processes tokens and vectors refer to anything? Yet the model's linguistic behavior continually invites precisely such attributions. The result is an unstable oscillation between dismissal and anthropomorphism. Either the model is said to manipulate only empty form, or it is suspected of possessing something like human understanding. Both reactions are misleading. The first ignores the internal organization produced by training on human language. The second forgets that the system remains a numerical mechanism rather than a human subject.

This pressure has a long history. Classical artificial intelligence was often understood through the paradigm of symbolic representation and rule-governed manipulation. Newell and Simon's physical symbol system hypothesis gave this tradition one of its most influential formulations: intelligent activity was to be explained in terms of symbol structures and processes operating over them \citep{Newell1976}. But this symbolic picture immediately raised a deeper question. If a system manipulates symbols only in virtue of their formal properties, what makes those symbols mean anything? Searle's Chinese Room argument sharpened this worry by insisting that syntactic manipulation alone does not amount to understanding \citep{Searle1980}. Harnad later formulated the issue as the symbol grounding problem: how can the semantic interpretation of a formal symbol system be made intrinsic to the system rather than merely parasitic on meanings supplied by external interpreters \citep{Harnad1990}?

The same problem returns in a new form for LLMs. Modern language models are not classical symbolic systems in the old sense. They do not manipulate hand-coded symbols by explicit rules. They compute over high-dimensional vectors, learned weights, activation patterns, attention operations, and output logits. Nevertheless, the old worry does not disappear. It becomes sharper. If a symbolic system may be accused of manipulating meaningless symbols, a language model may be accused of manipulating meaningless forms or vectors. This is why the form/meaning debate around LLMs is not a side issue. Bender and Koller argue that systems trained only on linguistic form lack the relevant connection between form and communicative meaning \citep{Bender2020}. The ``stochastic parrots'' critique gives this worry its most memorable expression: LLMs can produce fluent text by recombining linguistic forms in ways that may remain disconnected from understanding and meaning \citep{Bender2021}. Whether one accepts that critique or not, it correctly identifies the central philosophical pressure: linguistic fluency alone does not settle the question of meaning or reference.

The vector grounding problem is the contemporary form of this pressure. If LLMs compute over vectors rather than classical symbols, then the grounding question must be reformulated: how, if at all, can vector states and vector-mediated outputs be about entities, properties, events, or states of affairs in the world? Coelho Mollo and Milli\`ere argue that this is not simply the old symbol grounding problem repeated without change. The internal states of LLMs are vectorial, distributed, and shaped by training rather than by explicit symbolic interpretation \citep{Mollo2026}. They therefore distinguish several notions of grounding, including referential, sensorimotor, relational, communicative, and epistemic grounding. This distinction is useful, but for the present paper the most important relation is between referential and relational grounding. Referential grounding asks whether model states and outputs can be connected to entities, properties, or states of affairs beyond the model. Relational grounding, by contrast, concerns the structure of relations among linguistic or internal representations. These should not be treated as wholly independent issues in the case of LLMs. If LLM reference is possible at all, it will not arise by bypassing relational structure. It will arise, if at all, because human referential practice has left relational traces in language, and because training can transform those traces into stable numerical relations inside the model.

This is why grounding and reference cannot be separated. To ask whether an LLM is grounded is, at least in part, to ask whether its words, vectors, or outputs can refer. 
The skeptical answer says no: the model receives only linguistic form and therefore lacks genuine word-to-world connection. Externalist approaches resist this conclusion. Mandelkern and Linzen argue that LLM inputs are not bare strings but strings with natural histories of referential use \citep{Mandelkern2024}. Koch develops a stronger Kripkean version of this strategy, arguing that LLM-generated uses of names and related referential expressions can inherit reference through reference-sustaining chains of linguistic use \citep{Kripke1980,Koch2025}. On this view, the decisive point is not whether the model has directly perceived the referent, but whether its use of a term is appropriately connected to an existing historical practice in which that term has already been referentially fixed and transmitted. These responses are important because they show that reference need not require direct perception by each speaker or system. 
A human being can refer to Napoleon, electrons, black holes, or Peano without having directly perceived them. Reference can be transmitted through testimony, education, texts, scientific practices, and communal chains of use.

My argument does not reject this externalist insight. It extends it. A reference-sustaining chain may explain how an expression remains historically connected to a referent, but it does not by itself explain how the many profiles associated with that expression are selected, suppressed, combined, and reactivated in context. That is the task of the profile-based and vector-level account developed in this paper.

This externalist point should be pushed further. If reference can be transmitted through language, then linguistic form cannot be treated as a mere surface detached from reference. But it also follows that reference itself should not be modeled primarily as a private inner attachment between a word and an object. I may have first-person access to my own memories, images, associations, and acts of understanding. I do not have comparable access to the inner reference profiles of other speakers. When I judge that another speaker refers to Paris, electrons, Napoleon, or Peano, I do so on the basis of public performance: the speaker's ability to use words, answer questions, make distinctions, accept corrections, explain relations, and coordinate discourse with others. Human reference in others is therefore not known by inspecting private mental contents. It is attributed through patterns of linguistic and practical coordination.

This observation explains why a profile-based account of reference is needed. From the first-person point of view, a word-object picture can appear natural: a word is heard, and an image, memory, or thought may arise. But this cannot be the public basis of reference. What is publicly available is not a single private image or a single mental hook, but a structured pattern of use, reidentification, distinction, correction, inference, and continuation. Reference, as it functions in communication, is therefore not merely a local link between an isolated expression and an isolated object. It is a context-sensitive profile through which speakers coordinate around entities, situations, relations, and domains.

I accept the importance of the externalist and vector-grounding turn, but I argue that it remains incomplete in two respects. First, it requires a richer account of reference. The metaphor of a word or representation ``hooking onto'' the world captures something important, but it also encourages an overly atomistic picture of reference. It treats the central case as a link between an isolated expression and an isolated worldly item. This paper argues instead that reference is often profile-based and discourse-sensitive: expressions acquire their referential roles within broader patterns of use, distinction, correction, inference, and continuation.

Second, the debate requires a clearer account of numerical realization. If LLMs inherit referential structure from human language, how does that structure become usable inside a trained vector system? It is not enough to say that a vector state may be causally or informationally connected to the world through text. A vector grounding problem requires a vector-level account of grounding. One must explain how language-mediated referential relations become embeddings, weights, activation patterns, attention-mediated hidden states, output alignments, and logits. Without this mathematical step, the causal-informational relation remains philosophically suggestive but mechanistically underdescribed.

The central thesis of this paper is therefore twofold. Philosophically, reference should be understood not as a single fixed mental item or word-object hook, but as a referential profile. A referential profile is a structured set of descriptions, memories, perceptions, inferential routes, affective orientations, social uses, and correction practices through which an expression, sentence, or discourse can converge on an entity, situation, relation, or domain. Human beings realize such profiles through perception, episodic memory, embodiment, affect, and social responsibility. LLMs do not. But the absence of the human mode of reference does not by itself show that every derivative form of reference is absent.

Computationally, I argue that any reference available to an LLM must be derivative, language-mediated, and numerically structured. The model does not first possess meanings and then calculate with them. Rather, linguistic relations inherited from human practice are converted into numerical relations through optimization. Worldly entities and events shape human world-directed practice; human practice shapes language; language shapes training data; optimization shapes parameters; parameters shape hidden states, logits, and outputs. The relevant causal chain is not direct model perception, but mediated causal-informational inheritance. This chain does not establish human understanding. It shows how world-involving linguistic practice can become a source of numerical structure inside the model.

The argument that follows develops this position by connecting three levels that are often kept apart. At the philosophical level, reference must be expanded from a word-object hook to a profile-based and discourse-sensitive structure. At the computational level, this profile must be explained as a distributed numerical structure stabilized through training and reactivated through context-sensitive computation. At the evidential level, mechanistic interpretability findings can be read not merely as isolated curiosities, but as evidence that entity profiles, factual-referential relations, and affective-pragmatic orientations can become stable and causally active inside trained models.

The aim is not to show that LLMs refer as humans refer. They do not. Nor is it to claim that mathematics creates reference from nothing. It does not. The source of grounding remains human world-directed practice, sedimented in language. The claim is more limited but, I think, philosophically important: once such practice has entered linguistic form, optimization can transform its recurrent relations into stable, distributed, and causally active numerical structures. That is the sense in which derivative, language-mediated LLM reference, if it exists, is genuinely vector grounding.

\section{Reference, Grounding, and Referential Atomism}
The question whether LLMs refer has usually been framed against two opposing intuitions. On the skeptical side, LLMs are trained on linguistic form rather than on direct perceptual or practical interaction with the world. Bender and Koller define form as any observable realization of language---marks, pixels, bytes, or articulatory movements---and meaning as a relation between such form and something external to language, especially communicative intent \citep{Bender2020}. From this perspective, a system trained only on form lacks the relevant connection between linguistic expressions and the non-linguistic world or intentions that make those expressions meaningful.

This skeptical position has force. A model that receives only token sequences does not see Paris, touch apples, remember childhood events, or coordinate attention with another speaker toward a shared object. If reference requires this kind of human world-involvement, then LLMs plainly do not refer. But the question is whether reference itself requires this human form of world-involvement, or whether human reference is only one way in which reference can be realized.

A different answer is offered by externalist approaches to reference. Mandelkern and Linzen argue that the appearance that LLMs cannot refer is misleading because their inputs are not bare strings. They are strings with natural histories of referential use \citep{Mandelkern2024}. On this view, reference is not grounded primarily in an individual speaker's beliefs, experiences, or discriminatory capacities, but in causal-historical links between a linguistic community's use of a word and its referent. Koch develops a related Kripkean strategy, arguing that LLM-generated texts can meet the conditions for successful reference, at least for proper names and the so-called paradigm terms, by inheriting reference from training data through a reference-sustaining mechanism \citep{Koch2025}.

These externalist responses are important because they loosen the tie between reference and direct acquaintance. A human speaker can refer to Peano, Napoleon, electrons, or black holes without having perceived them directly. In many such cases, direct acquaintance is unavailable, historically impossible, or scientifically mediated. 
Reference can nevertheless be transmitted through testimony, education, books, scientific practice, and communal chains of use. If so, the absence of direct perception cannot by itself settle the question of LLM reference.

But externalism also leaves something underdescribed. To say that LLMs inherit natural histories of referential use from training data explains how their words may be connected to prior human uses. It does not yet explain what kind of internal organization allows such inherited reference to guide the model's behavior. Nor does it fully explain what reference is, if it is not to be reduced either to direct acquaintance or to a bare causal chain of word use. The appeal to natural history is powerful, but it risks making reference look too thin: as if reference were simply a historical line running from an output token back through training data to a human use and finally to an object.

The vector grounding approach sharpens this issue. Coelho Mollo and Milli\`ere argue that modern LLMs should not be treated as classical symbolic systems. They compute over vectors, and therefore the relevant question is how vector states and vector-mediated outputs can be about extra-linguistic reality \citep{Mollo2026}. They distinguish several notions of grounding that are often conflated: referential grounding, sensorimotor grounding, relational grounding, communicative grounding, and epistemic grounding.

Referential grounding is the central notion for their purposes. It concerns how a representation connects to a worldly referent. They sometimes describe this as the problem of how representations ``hook onto'' things in the world \citep{Mollo2026}. Sensorimotor grounding, by contrast, concerns either the connection between conceptual or linguistic representations and sensorimotor representations, or the direct contact between sensorimotor transducers and the world. Relational grounding concerns intra-linguistic relations, such as the way a word's meaning may be partly determined by its relation to other words. Communicative grounding concerns the coordination of understanding between interlocutors in conversation. Epistemic grounding concerns the connection between linguistic representations and stored knowledge.

The significance of this distinction is that Coelho Mollo and Milli\`ere reject a simple sensorimotor requirement. Direct causal contact with the world is not, on their view, necessary for referential grounding. Nor is relational grounding by itself sufficient, since word-to-word relations alone leave unanswered how any representation connects to extra-linguistic reality. They therefore argue that referential grounding requires two teleosemantic conditions: internal states must stand in appropriate causal-informational relations to the world, and they must have a selection history that gives them the function of carrying that information \citep{Mollo2026}.

I agree with this rejection of both simple sensorimotor grounding and purely intra-linguistic grounding. But in the case of LLMs, the relation between referential and relational grounding must be stated more carefully. Relational structure is not a substitute for referential grounding, but neither is it a merely secondary phenomenon. If LLMs can acquire any derivative form of referential grounding, they can do so only because human referential practice has left relational traces in language, and because training transforms those traces into stable numerical relations inside the model. In this sense, relational structure is not a rival to referential grounding. It is the medium through which mediated, machine-specific referential grounding could become available to a vector system.

This is a valuable advance. It avoids the crude claim that LLMs must be embodied or multimodal in order to refer. It also avoids the opposite claim that intra-linguistic relations alone are enough. But even this more sophisticated account remains shaped by what may be called \emph{referential atomism}. By referential atomism I mean the tendency to treat the basic case of reference as a relation between an isolated expression and an isolated worldly item: a name and a person, a word and an object, a representation and its referent. The ``hooking onto'' metaphor is not responsible for this tendency by itself, but it reinforces it. It encourages us to imagine reference as a line running from a linguistic item to a worldly entity.

This picture is not simply false. Proper names, natural kind terms, and demonstratives often make word-object reference especially salient. A theory of reference must explain how ``Paris,'' ``Napoleon,'' or ``electron'' can be about Paris, Napoleon, or electrons. But reference in ordinary language rarely operates only at this atomic level. Words function inside sentences; sentences function inside discourse; discourse establishes situations, develops lines of thought, introduces contrasts, corrects misunderstandings, and guides hearers toward shared structures. In such cases, reference is not merely a local attachment between one word and one object. It is part of a larger organization of context.

The limitation of referential atomism becomes especially clear in extended discourse. An explanation, argument, or instruction is not understood by identifying the object named by each word one by one. It may introduce entities, properties, relations, idealizations, contrasts, and inferential roles that must be followed as a whole. The relevant reference is therefore not only word-object reference, but discourse-level orientation: the stabilization of a shared line of thought through which objects, properties, relations, and inferential roles become jointly organized.

This point is especially important for LLMs. A transformer does not process each token as an isolated word-object hook. A token representation is repeatedly updated by context. The representation associated with ``Paris'' in a question about capitals is not the same, functionally, as the representation associated with ``Paris'' in a sentence about the Louvre, the 2015 attacks, fashion, or Paris, Texas. The relevant issue is therefore not whether the token ``Paris'' is attached once and for all to a single object. The issue is whether the model can use context to activate the appropriate referential profile.

The problem, then, is not that previous accounts are wrong. The problem is that each captures only part of the phenomenon. The form-based skeptic rightly insists that human meaning involves more than surface strings. The externalist rightly insists that reference need not depend on private beliefs or direct acquaintance. The vector grounding theorist rightly insists that grounding requires causal-informational and historical relations to the world. But all three approaches tend to leave underdeveloped the discourse-level and profile-based character of reference. They ask how an expression is connected to a referent, but they say less about how a whole linguistic context selects, maintains, and corrects the relevant referential profile.

What remains missing is therefore a richer account of reference itself: one that
can explain why human reference is not a single fixed mental item, why reference
can be sustained without direct acquaintance, why word-to-word relations in human
language are not merely empty formal relations, and why machine reference, if it
exists, need not take the same form as human reference. 

\section{Human Reference as Profile-Based and Context-Sensitive}
The preceding section identified a recurring limitation in current debates about LLM reference. Skeptical, externalist, and vector-grounding approaches disagree about whether LLMs can refer, but they often leave reference itself thinner than it should be. The skeptical view tends to measure reference against human perception, memory, and communicative intention. The externalist view emphasizes natural history and causal chains of linguistic use. The vector grounding view emphasizes causal-informational relations and selection history. Each of these captures something important. None by itself captures the full structure of reference as it appears in ordinary human language.

Consider again the word ``Paris.'' A speaker who hears this word may think of the Eiffel Tower, the Seine, the Louvre, the French capital, a childhood trip, a political protest, the 2024 Olympics, a map, a smell, a film scene, or a historical event. Another speaker may activate a different set of associations. Even the same speaker may activate different aspects in different contexts. In a geography lesson, Paris may be understood primarily as the capital of France. In an art-historical context, it may evoke museums and artistic movements. In a travel conversation, it may evoke tourism and personal memory. In a security discussion, it may evoke terrorist attacks or political events.

This variability does not show that ``Paris'' lacks reference. Nor does it show that each use refers to a different object. Rather, it shows that reference is not identical with any one mental image, memory, definition, or sensory trace. The reference of ``Paris'' is stabilized through a profile: a structured set of routes by which speakers can identify, reidentify, describe, distinguish, correct, and make true or false claims about the city. Some routes are perceptual, some testimonial, some historical, some affective, some inferential, and some purely linguistic. The referent is not any one of these routes. Reference is the organized convergence of these routes on an entity or domain.

By a \emph{referential profile}, I therefore do not mean a private mental image or a list of subjective associations. I mean a structured set of discriminative, inferential, corrective, affective, and pragmatic dispositions through which an expression or discourse can be directed toward an entity, situation, relation, or domain. A profile determines what counts as the same object across contexts, what must be distinguished from it, what claims about it are correct or incorrect, what inferences it licenses, what emotional or practical orientations may become salient, and what continuations are appropriate in discourse. It is therefore not an inner picture attached to a word, but a context-sensitive organization of possible uses, corrections, contrasts, and continuations.

This profile-based account clarifies why human reference is both stable and variable. It is stable because the different routes can be corrected and coordinated. If someone says that Paris is the capital of Germany, the claim can be corrected. If someone confuses Paris, France with Paris, Texas, the distinction can be made. If someone associates Paris only with the Eiffel Tower, further descriptions can expand the profile. But reference is also variable because no speaker activates the whole profile at once. Different aspects become salient depending on context, memory, affect, and purpose.

The same point applies to extended discourse. In a physics lesson, students may follow the same explanation through different routes: one visualizes a diagram, another follows the algebra, another recalls an experiment, and another retains only the general physical situation. Their images, memories, emotions, and inferential paths differ, yet successful communication does not require identical inner representations. It requires sufficient convergence on the same quantities, relations, contrasts, and inferential roles.

Yet this difference does not normally destroy communication. If the explanation succeeds, the students do not have identical inner representations, but they nevertheless converge on a common structure. They may differ in what they imagine, remember, or emphasize, but they can still follow the same broad physical situation, identify the same quantities, distinguish the same alternatives, and answer related questions in sufficiently similar ways. Human communication therefore does not require identical reference profiles. It requires enough overlap, coordination, and correction among different profiles for a shared line of thought to be maintained.

This point is crucial. Human reference is not private mental duplication. A speaker does not transmit an inner image into another mind. Rather, linguistic communication works by guiding different hearers, with different memories and backgrounds, toward a sufficiently common referential organization. The stability of reference lies not in the sameness of internal experience, but in the ability of diverse profiles to converge on common entities, relations, contrasts, and inferential roles.

This emphasis on public coordination is continuous with familiar themes in later Wittgenstein and semantic externalism. Meaning and reference are not fixed by private inner items alone, but by patterns of use, correction, and socially sustained relations to the world \citep{Wittgenstein1953,Putnam1975,Burge1979}. The present account extends this point by treating reference as a profile that can be publicly coordinated among humans and, in a derivative form, numerically stabilized in LLMs.

This also explains why reference is norm-governed. A referential profile is not a free association of impressions around a word, but a structure within which some continuations count as appropriate, others as mistaken, and corrections as intelligible. Reference is therefore not merely associative. It is a field of possible agreement, disagreement, correction, and continuation.

The relevance to LLMs is direct. LLMs do not possess the human profiles just described. They do not form visual images, recall personal episodes, feel boredom, understand a lecture as a student does, or become excited by a new idea. They do not share the human mode of reference. But linguistic data contain the traces of countless acts of human communication in which different speakers and hearers, with different reference profiles, nevertheless converge on common topics, entities, events, relations, and lines of inference. What is repeated across such communicative situations is not a private mental image, but a publicly stabilized pattern of use, correction, distinction, and continuation.

This is the level at which LLMs can enter the discussion. A model trained on language does not acquire the teacher's experience, the student's insight, or the listener's memory. But it can be exposed to the recurrent linguistic structures through which such experiences, insights, memories, corrections, and explanations are coordinated in human discourse. The common denominator among diverse human reference profiles becomes available in language as a pattern of relations. A trained model may parameterize this common structure without sharing the human sensory or episodic basis from which it originally arose.

The contrast should therefore not be between humans who possess one fixed reference and LLMs that possess none. Human reference itself is not fixed in that way. The relevant contrast is between different modes of profile formation. Human referential profiles are formed through perception, memory, embodiment, affective life, testimony, correction, and social responsibility. If they possess any referential organization at all, it must be derivative: inherited from publicly stabilized patterns of human linguistic practice rather than generated from first-person perceptual or episodic contact with the world.

This reformulation also explains why linguistic context becomes philosophically central. A single expression can participate in many profiles, and no isolated occurrence is enough to determine how it should be used, contrasted, corrected, or continued.

\section{Linguistic Form Is Not Empty Form}
A common skeptical claim is that systems trained only on linguistic form cannot acquire meaning. In one influential formulation, Bender and Koller argue that ``a system trained only on form has a priori no way to learn meaning,'' where meaning is understood as a relation between linguistic form and communicative intent \citep{Bender2020}. This claim has force if the target is human-analogous understanding: a system that receives only linguistic forms does not thereby acquire a speaker's intentions, perceptual situation, embodied activity, or responsibility in communication. In that sense, an LLM does not understand language as a human interlocutor understands it.

But this does not settle the question of reference. The skeptical inference becomes too quick if it moves from the absence of human communicative understanding to the absence of any machine-specific referential organization. The phrase ``trained only on form'' can suggest that the training material is merely a collection of uninterpreted marks. If form means random signs wholly detached from world-directed use, then the conclusion is right: a model trained on such marks would not thereby acquire meaning or reference. 
But ordinary linguistic form is not normally like this. It is produced and interpreted within human practices of perceiving, acting, measuring, remembering, correcting, testifying, and referring.

The case of a book helps to clarify the issue, but only if the analogy is kept limited. 
A book is physically an arrangement of marks on paper or pixels on a screen. The book itself does not perceive, intend, remember, understand, or refer as a human speaker does. We do not normally ask whether the book understands what it says, because we treat it as a static vehicle of language whose words were produced by a human author. In reading it, we presuppose that the text belongs to a human linguistic practice and that its marks are connected, through authorship, interpretation, and use, to human ways of perceiving, acting, remembering, correcting, and referring. The significance of the marks is therefore not generated by the physical object itself, but neither are the marks mere scratches.

The analogy with LLMs begins here, but it also ends here. An LLM response may look, at the surface, like written text, but it is not simply a fixed inscription left by a human author. It is produced interactively by a trained numerical system in response to a context. This is why the question arises for LLMs in a way it does not arise for books. We do not need to ask whether a book understands, because the relevant human source is already presupposed. With an LLM, that presupposition is unavailable: the system generates new responses without possessing human perception, memory, intention, or responsibility. The question is therefore not whether the LLM is just like a book, nor whether it has human understanding, but whether the human referential practices sedimented in language can be transformed into a machine-specific numerical structure capable of stabilizing and reactivating referential profiles.

The book example shows why the contrast between form and meaning must be handled carefully. If form is considered in complete isolation from any practice of use or interpretation, then it is empty. But human linguistic form is not normally encountered in this way. It is produced by writers, speakers, scientists, witnesses, teachers, officials, historians, and ordinary language users who are embedded in the world. It is then interpreted by readers and hearers who bring their own memories, expectations, background knowledge, and practical concerns to the text. Meaning is not located in the ink as a mental state. But neither is the ink merely meaningless once it is part of a linguistic practice. The physical form is inert; the linguistic form is not empty.

This point can be put in a more general epistemological form. I may have direct access, if anything, only to my own memories, images, associations, and acts of understanding. I do not have such access to the inner reference profiles of other speakers. When I judge that another speaker refers to Paris, electrons, Napoleon, or Peano, I do so on the basis of public performance: the speaker's ability to use words, answer questions, make distinctions, accept corrections, explain relations, and coordinate discourse with others. Human reference in others is therefore not known by inspecting private mental contents. It is attributed through patterns of linguistic and practical coordination.

This also explains why a profile-based account is not optional. From the first-person point of view, a word-object picture may seem natural: I hear a word, and an image, memory, or thought may arise in me. But this cannot be the public basis of reference. I cannot inspect whether another speaker has the same image, memory, or thought. What can be publicly tested is whether the speaker can use the expression across contexts, distinguish nearby cases, accept correction, paraphrase the same point, answer relevant questions, and maintain a shared line of discourse. Public reference therefore cannot depend on a single private hook between word and object. It must depend on a profile: a structured pattern of use, reidentification, contrast, correction, inference, and continuation through which speakers coordinate around a common topic or domain.

This observation matters because it shows that reference is already publicly stabilized in language. If testimony, books, lectures, and conversations can transmit reference among humans who lack direct acquaintance with many of the things they discuss, then linguistic form cannot be treated as empty marks detached from reference. A person can refer to Napoleon without having met Napoleon, to electrons without having seen an electron, and to black holes without having perceived one directly. Such reference is possible because language carries stabilized traces of prior world-directed practice. The relevant question for LLMs is therefore not whether text, considered as bare uninterpreted form, magically creates reference. It does not. The question is whether a model trained on language can transform these publicly stabilized referential profiles into machine-usable numerical structures. In a finite vector system, such profiles cannot appear as one word, one image, one neuron, or one coordinate. They must appear as distributed, context-sensitive patterns of activation, weighting, alignment, and continuation.

This point matters for LLMs. It is tempting to say that an LLM trained only on form is like a book: a system of marks without understanding. There is an important truth in this comparison, but the comparison must be handled carefully. In the case of a book, the absence of understanding in the physical object is not troubling, because the text is treated as the product of a human author. A reader may interpret the marks as meaningful because they are presumed to originate in human linguistic practice, human intention, and human reference. The main question is therefore not whether the book understands what it says. It plainly does not.

The LLM case is different. The problem does not arise primarily on the reader's side: a human reader can interpret an LLM output as meaningful, just as she can interpret a sentence in a book as meaningful. The problem arises on the production side. An LLM generates new responses without being a human author and without possessing human perception, memory, intention, or responsibility. The question is therefore not whether a reader can attach meaning to the output, but whether the process that produces the output contains any machine-specific form of referential organization. If the model has no human understanding, that does not by itself show that its outputs are empty marks. It shows that whatever referential organization it has must be derivative, language-mediated, and numerically realized in the trained system.

But the comparison with a book also reveals why the skeptical conclusion is too quick. The fact that a book does not understand does not imply that its linguistic form is empty. Likewise, the fact that an LLM does not possess human understanding does not imply that the linguistic form on which it is trained is meaningless material. The relevant question is not whether the model has the mental life of a reader. The question is whether training on human linguistic form can stabilize patterns that are already shaped by human referential practice.

There is also a crucial difference between a book and an LLM. A book is a static inscription. Once written, it presents the same sequence of words to different readers, although different readers may interpret it differently. An LLM is not static in this way. It is a trained generative system that produces new linguistic forms in response to input. It does not merely display a fixed text. It selects, combines, and continues linguistic patterns according to the current context. For this reason, the question of reference arises differently for an LLM than for a book. We need not ask whether the model understands as a human reader understands. But we can ask whether its context-sensitive generation is guided by referential profiles inherited from human language.

This is especially clear in the case of reference. ``Paris'' does not occur randomly in human language. It occurs with ``France,'' ``capital,'' ``Eiffel Tower,'' ``Seine,'' ``Louvre,'' ``tourism,'' ``revolution,'' ``fashion,'' ``Olympics,'' ``map,'' ``airport,'' and many other expressions. These relations are not arbitrary formal coincidences. They are linguistic traces of how humans have encountered, described, mapped, visited, governed, remembered, and argued about Paris. The model does not see Paris. But the linguistic form on which it is trained is already shaped by human relations to Paris.

The crucial point is that a single expression can participate in many different referential profiles. Depending on context, ``Paris'' may activate a capital-city profile, a cultural-geographical profile, a historical-political profile, or a contrastive profile such as Paris, Texas. These are not arbitrary meanings attached to the same string, but context-sensitive ways in which one expression can be directed toward different aspects, relations, contrasts, or domains.

This is why large-scale linguistic data matter. The point is not that the model needs many examples from which to estimate frequencies. The point is that human referential practice leaves its traces across many contexts of use, and those traces must recur often enough, and in sufficiently varied settings, to become stable under optimization. A single occurrence of a relation is not enough to form a durable numerical profile. Even a simple profile such as Paris as the capital of France must appear repeatedly across descriptions, questions, corrections, comparisons, explanations, and continuations before it can exert a stable directional pressure on the parameter system. Large-scale data are therefore required not only because one expression can participate in many profiles, but also because each profile must be repeatedly encountered across different contexts in order to become numerically stabilized. Perception, memory, affect, correction, testimony, explanation, and practical involvement do not enter the model as first-person experience. They enter, if at all, as linguistic residues distributed across many contexts. For LLMs, the non-linguistic supports of human reference can become available only through these repeated linguistic traces.

Thus the slogan ``trained only on form'' is misleading if it treats form as independent of the practices that produced it. The relevant contrast is not between world and form, as if form were simply outside the world. The contrast is between direct human world-involvement and mediated linguistic inheritance. LLMs do not acquire reference by seeing, touching, remembering, or participating in shared practical situations. They acquire, at best, a derivative form of reference by being trained on the linguistic residues of human reference. But those residues are not empty. They contain patterns of use, correction, contrast, explanation, and continuation through which human speakers have stabilized referential profiles.

This point also changes how next-token prediction should be understood. It is true that the surface task of an LLM is to predict or select a continuation. But successful continuation often requires more than local word association. In the context ``Paris is the capital of,'' the relevant continuation depends on activating a France-capital profile. In the context ``The Louvre is located in,'' the relevant continuation depends on activating a cultural-geographical profile. In the context ``the 2015 attacks in Paris,'' the relevant continuation depends on a historical-event profile. In the context ``Paris, Texas,'' the model must avoid the France-capital profile and select a different place-name profile. If the wrong profile is activated, the continuation may be irrelevant or false.

Successful continuation therefore requires context-sensitive referential profile selection. This does not mean that the model possesses human reference. It means that linguistic performance itself depends on structures that are reference-like in a derivative and language-mediated sense. A model that consistently distinguishes Paris, France from Paris, Texas, connects the Louvre with Paris but not with Berlin, and treats ``capital of France'' differently from ``site of the Louvre'' is not merely responding to an isolated string. Its behavior is guided by patterns inherited from human referential practice.

\section{From Linguistic Form to Numerical Parameterization}
The vocabulary of ``learning'' can obscure the distinction at stake. The problem is not merely terminological. To say that a machine ``learns'' easily invites an anthropomorphic picture, as if the system acquired understanding in a manner continuous with human learning. This tendency is difficult to avoid even in ordinary interaction with LLMs. Because they respond fluently, correct mistakes, continue arguments, and adapt to conversational context, users naturally address them as if they were interlocutors. One may become irritated, disappointed, persuaded, or surprised by their outputs, even while knowing that there is no human subject on the other side. The language of learning strengthens this tendency.

The deeper point is that an LLM, as software running on a computer, can do only what computers do: computation. It does not operate on words as humanly meaningful words. It operates on numerical representations. The fact that the input and output appear to human users as language can hide this point. A user types or speaks words; the system returns words. From the outside, the process looks like linguistic exchange. But internally the process requires two translations. The input must first be converted into tokens and numerical vectors. After computation, the numerical output must be converted back into token identifiers and then into human-readable text.

This double translation is familiar from scientific computation. A band-structure code, for example, does not directly produce physical understanding. It receives input in a prescribed formal format, performs numerical operations, and outputs quantities such as eigenvalues at different \(k\)-points. Those numbers become physically meaningful only within a surrounding practice of interpretation: one may plot them as a band diagram, compare gaps, identify dispersions, or relate the result to a material system. The software does not understand the band structure. It computes. The physicist interprets the numerical result within a theoretical and practical framework.

The case of LLMs is different in surface appearance but similar in this basic respect. The model does not output meaning directly. It computes over numerical structures and produces a distribution over possible next tokens. A simplified way to state this is that the system produces scores, or logits, for vocabulary items. A decoding procedure then selects or samples token identifiers, which are finally rendered as written words. Because the final result appears as ordinary language, it is easy to forget that the system has not left the domain of computation. The apparent linguistic immediacy of the output conceals the numerical mediation that makes it possible.

For this reason, I avoid using ``learning'' for LLMs except when referring to the conventional expression ``machine learning.'' Even the term ``training'' is not entirely neutral, since it too can suggest the training of an agent or learner. I use it only as a technical shorthand. What is at issue is not human learning, but numerical parameterization under optimization. Human learning involves perception, memory, embodiment, social correction, affective response, and the reorganization of understanding within a lived world. LLM optimization, by contrast, adjusts a parameter system under an objective function. More specifically, it is a form of relational parameterization: recurrent relations in linguistic data are stabilized in embeddings, weights, activation patterns, and output dispositions so that they can be computationally reused.

This point is crucial for the problem of reference. A computer cannot calculate a word as a word in the human sense. It can calculate numerical relations among token representations. Thus, if an LLM has any derivative form of reference, it cannot consist in a human-like act of attaching a word to an object. It must consist in the numerical stabilization of relations among tokens, contexts, and continuations that have been shaped by human referential practice. The model does not first possess meanings and then calculate with them. Rather, whatever meaning-like or reference-like structure it has must arise from the way linguistic relations are converted into numerical relations and reused in context.

This also explains why the final textual output can be misleading. When a model answers in a sentence, the user sees language and naturally evaluates it as language. This is appropriate at the level of interpretation. But at the level of mechanism, the sentence is the result of numerical operations over token representations. The model has not produced a thought and then expressed it in words. It has transformed an input sequence into internal vectors, propagated those vectors through learned transformations, produced token scores, and rendered selected tokens as text. The philosophical question is therefore not whether a hidden human-like speaker stands behind the sentence. The question is whether the numerical process can preserve, select, and reactivate referential profiles inherited from human linguistic form.

This formulation avoids two opposite mistakes. The first mistake is anthropomorphism: treating the model as if it were a human learner that understands its words as we do. The second mistake is empty formalism: treating the model's outputs as if they were mere strings with no internal organization connected to human referential practice. The right intermediate position is that an LLM is a numerical system whose computations are constrained by parameterized traces of human language use. Its outputs are not human thoughts. But neither are they arbitrary marks. They are generated from numerical structures shaped by the relations among words, contexts, corrections, and continuations in human linguistic practice.

This terminological caution matters for the problem of reference. The question ``Does the model understand?'' or ``Does the model know what its words refer to?'' easily imports the wrong standard. It treats the model as if it were a deficient human speaker: a would-be subject lacking perception, memory, intention, and responsibility. But the question of LLM reference should not be framed as whether a machine has human understanding hidden inside it. The relevant question is whether the numerical system produced by optimization can inherit, stabilize, and reactivate referential profiles sedimented in human linguistic practice.

This also clarifies the analogy with books. A book does not understand the sentences printed in it. It has no beliefs, perceptions, intentions, or memories. Yet we do not conclude that the linguistic forms in a book are meaningless scratches. Their significance depends on the human practices in which they were written, read, interpreted, corrected, and reused. An LLM is not a book, because it is not a static inscription: it generates new linguistic forms in response to input. But neither is it a human learner. It is a trained numerical system whose outputs are shaped by parameterized traces of human linguistic practice.

Thus the issue is not whether the model ``learns'' in the human sense. It does not. Nor is the issue whether optimization alone creates reference from nothing. It does not. The issue is whether optimization can transform the referential profiles already present in human linguistic form into a numerical structure that can guide context-sensitive continuation. In this limited but important sense, the relevant process is better described as relational parameterization than as learning.

This is neither human understanding nor frequency bookkeeping. The model is not given a sampled survey of language from which it merely estimates verbal probabilities. It is exposed to large bodies of linguistic material in which human referential practice has already left structured traces. Optimization then adjusts a high-dimensional parameter system so that recurrent relations in that material become computationally usable. The familiar surface effects of LLM output---generic phrasing, conventional continuations, or overused expressions---can make the process look like aggregate verbal averaging. But those effects should not be mistaken for the whole mechanism. An appropriate continuation is produced only when the current context activates and aligns with a relevant profile among many competing profiles. What matters for the present argument is therefore not statistical counting, but the numerical stabilization of relational structure.

The previous section argued that a single expression can participate in multiple referential profiles. This makes the training problem more demanding than attaching one word to one object. The model must become sensitive to the contexts in which different profiles are relevant. It must be trained that ``Paris'' participates in one profile when it appears with ``capital,'' ``France,'' and ``government,'' another when it appears with ``Louvre,'' ``Seine,'' and ``museum,'' another when it appears with ``attack,'' ``2015,'' and ``Bataclan,'' and still another when it appears with ``Texas.'' The relevant structure is not a list of definitions. It is a set of context-sensitive relations distributed across many expressions and uses.

The term ``causal-informational'' can be misleading if it is understood as requiring a direct stimulus-response relation between the world and the model. That is not the relation at issue here. An LLM does not stand to Paris as a perceptual system stands to a visible object. Paris does not directly stimulate the model's sensory surfaces, because the model has no such surfaces. The relevant causal relation is mediated and historical. Worldly entities and events affect human perception, memory, measurement, action, and testimony; these activities shape linguistic production; linguistic production forms the training data \(D\); and \(D\), through optimization, alters the parameters \(\theta\). In this sense, the data are a causal source of the model's numerical structure.

The relation is informational because the resulting numerical structure can preserve distinctions that originate in human world-directed practice. For example, the distribution of contexts in which ``Paris'' occurs carries information about Paris as the capital of France, the city of the Louvre, a tourist destination, a site of historical events, and a name to be distinguished from Paris, Texas. If optimization stabilizes these distinctions in the model's parameters, then later hidden states and logits can be guided by information inherited from human referential practice. The model has not perceived Paris. But its parameters may encode traces of how humans have referred to Paris.

Thus the causal-informational relation relevant to LLMs is not:
\[
\text{world} \rightarrow \text{model perception}.
\]
It is:
\[
\text{world} \rightarrow \text{human world-directed practice}\rightarrow \text{language} \rightarrow D \rightarrow \theta \rightarrow h \rightarrow z_v .
\]
This chain does not establish human understanding. It shows how world-involving linguistic practice can become a source of numerical structure inside the model. The remaining question is whether that numerical structure can be used to select and maintain the appropriate referential profile in context.

Relational parameterization names this middle transformation. The model does not store human experiences of Paris. Rather, training distributes the recurrent relations among expressions, contexts, and continuations across embeddings, weights, and activation patterns. What becomes stabilized is not a human-readable concept, but a machine-usable numerical structure that can participate in later computation.

This point is essential for understanding inference. When an input is given, the model does not merely retrieve a fixed meaning attached to each token. The input context activates a pattern of internal states, and that pattern must select among the referential profiles made available by training. In different contexts, ``Paris'' may activate a France-capital profile, a cultural-geographical profile, a historical-event profile, or a Paris-Texas profile. The same token can therefore participate in different computational trajectories depending on context.

This is why reference is not an optional supplement to next-token prediction. Next-token prediction is the surface task; context-sensitive referential profile selection is one of the internal conditions for performing that task successfully. A model can produce an appropriate continuation only if the current hidden state is aligned with the relevant profile rather than with an irrelevant neighboring profile. If the wrong profile is activated, the model may generate a fluent but false or irrelevant answer. The familiar phenomenon of hallucination can often be understood in this way: not as the absence of structure, but as the activation, recombination, or continuation of an inappropriate structure, especially when the training data are incomplete, incorrect, biased, or otherwise misaligned.

This formulation preserves what is right in the vector grounding approach while making explicit what it leaves underdeveloped. The causal-informational relation does not arise from mathematics alone. Mathematics does not create reference from nothing. But whatever referential structure is inherited from human linguistic practice becomes available to an LLM only by being numerically structured. Training is the process by which the many profiles sedimented in language are transformed into parameterized relations. Inference is the process by which an input context reactivates and selects among those relations.

\section{The Numerical Structuring of Referential Profiles}
If reference is profile-based, then it should not be expected to appear inside a model as a single symbol, coordinate, neuron, or human-readable concept. A referential profile such as ``Paris'' or ``Golden Gate Bridge'' involves many relations: names, descriptions, locations, visual associations, factual links, contrast classes, affective tones, and pragmatic uses. In a finite model, such profiles cannot be stored as isolated containers. They must be distributed across many learned relations, and they may share representational resources with other profiles.

The preceding account of mediated causal-informational inheritance remains too abstract unless we specify how such relations become operative inside the model. It is not enough to say that human linguistic practice causally shapes the training data and that the data causally alter the parameters. We must also ask what kind of causal structure a parameter actually has within the computation. 
A minimal algebraic point helps clarify this.

Consider a vector \(x \in \mathbb{R}^d\), a weight matrix \(W\in \mathbb{R}^{d^\prime \times d}\), and an output vector \(y\in \mathbb{R}^{d^\prime}\), with
\[
y_i=\sum_j W_{ij}x_j.
\]
The \(i\)-th output coordinate is not produced by the entire matrix in an undifferentiated way. It is determined by the \(i\)-th row of \(W\):
\[
y_i = W_{i\cdot}x.
\]
Other rows do not directly determine \(y_i\). They determine other output coordinates. 
Thus each row defines a particular response profile: it specifies how the whole input pattern is to be read in order to produce one output coordinate. The input vector as a whole matters, but it matters through the row-specific pattern of weights. In differential form,
\[
\frac{\partial y_i}{\partial x_j}=W_{ij}.
\]
This equation gives a precise sense in which a trained weight is a local causal sensitivity within the model's computation. If \(x_j\) changes while the other components are held fixed, \(y_i\) changes in proportion to \(W_{ij}\). The weight is not a semantic label. It does not say, by itself, ``Paris,'' ``capital,'' or ``France.'' But it does determine how one numerical component can affect another. It is therefore one site at which a relation obtained from data becomes a computationally active causal coefficient.

The column view gives the complementary point. The \(j\)-th input coordinate contributes to the whole output vector through the \(j\)-th column:
\[
\frac{\partial y}{\partial x_j}=W_{\cdot j}.
\]
The column \(W_{\cdot j}\) specifies how a change in one input coordinate is distributed across all output coordinates. Holding the other input coordinates fixed, the direct local effect of \(x_j\) is given by this column alone; the other input columns do not enter this derivative. Thus an input coordinate does not have a single isolated effect. It has a profile of effects across the next representation. Rows describe how each output coordinate reads the whole input pattern; columns describe how each input coordinate propagates its influence into the next layer. Together, they show that the model's internal causality is both coordinate-specific and distributed.

This is the mathematical point needed for the grounding argument. A causal-informational relation inherited from human language does not enter the model as a single symbolic fact or as a direct word-object hook. During optimization, repeated relations in the data alter many such sensitivities \(W_{ij}\). A relation such as Paris--capital--France is not stored as one sentence or one coordinate. It is dispersed across many trained coefficients that determine how token representations, contextual features, and later activations influence one another. Training therefore converts recurrent linguistic relations into a structured field of numerical causal sensitivities.

The role of the softmax objective is important here. A next-token prediction is not trained as an isolated association between one context and one token. At the output layer, the model assigns logits to the vocabulary and the softmax converts these logits into a distribution:
\[
p(v\mid h)=\frac{\exp(z_v)}{\sum_{u\in V}\exp(z_u)}.
\]
With cross-entropy loss, the gradient with respect to each logit is
\[
\frac{\partial L}{\partial z_v}=p(v\mid h)-\mathbf{1}_{v=y},
\]
where \(y\) is the target token. Thus the target token is not adjusted in isolation. Its score is trained against the scores of all other candidate tokens. In a full-softmax setting, every prediction is therefore a contrastive event over the whole vocabulary: the context must increase the appropriateness of the target continuation while decreasing, to varying degrees, the appropriateness of competing continuations. This is one reason multiple referential profiles can be stabilized in a shared parameter space. A context involving ``Paris'' does not merely strengthen a connection between ``Paris'' and one next word; it adjusts the relative compatibility among many possible continuations---``France,'' ``Louvre,'' ``Texas,'' ``fashion,'' ``attack,'' and countless others. Repeated across many contexts, these contrastive adjustments distribute profile-relevant distinctions across embeddings, output weights, and internal transformations.

This also explains why the same computational structure matters again at inference. The model does not use one mechanism during optimization and a wholly different mechanism when answering. The parameters shaped during optimization are the parameters through which later inputs are processed. When a new context containing ``Paris'' is presented, the resulting vector propagates through sensitivities adjusted by prior exposure to Paris-related contexts. Depending on the context, this propagation may favor a France-capital profile, a cultural-geographical profile, a historical-event profile, or a Paris-Texas profile. In this way, causal-informational inheritance from language becomes operative as context-sensitive numerical propagation.

The philosophical point is therefore not merely that a matrix multiplication obeys a mathematical rule. It is that trained weights provide the exact places where inherited linguistic relations become computationally causal. The data influence the parameters during optimization; the parameters influence hidden states and logits during inference. This two-stage structure is what allows a mediated causal-informational relation to become a numerically stabilized referential profile inside the model. The strictness of the computation matters here. Once relations have been encoded in the parameters, they are not applied loosely, metaphorically, or intermittently. They are applied through the same mathematical operations each time the model processes an input. Unlike a human reasoner, who may forget, misapply, or inconsistently follow a rule, the machine executes the learned transformations with mechanical regularity. This is simply how software ordinarily operates: specified transformations are applied without discretion, hesitation, or forgetfulness. This regularity does not create reference by itself, but it is what allows inherited linguistic relations to be repeatedly reactivated, reinforced, and stabilized as usable numerical structure. What may look, at the surface, like statistical averaging is often the visible result of deterministic transformations being applied without exception to large bodies of structured linguistic data.

Superposition enters at this point, but it should not be overstated. In a finite model, many profiles must share representational resources. Paris, London, Rome, Seoul, city, capital, tourism, river, museum, and Olympics cannot each occupy fully isolated representational compartments. Their profiles overlap. Some coordinates and directions will participate in more than one structure. Superposition explains this sharing. Distribution explains recoverability. If several profiles share coordinates, then no single coordinate can simply be read as one referent. A profile becomes usable only as a pattern distributed across many coordinates, directions, and transformations.

Before the final output alignment occurs, however, the current hidden state must first be formed. This is where the transformer architecture matters. In an autoregressive transformer, token representations are not processed as isolated word types. At each layer, a token representation is updated in relation to other tokens in the available context through self-attention. In simplified form, attention computes
\[
\mathrm{Attention}(Q,K,V)
=
\mathrm{softmax}\!\left(\frac{QK^{\top}}{\sqrt{d_k}}\right)V,
\]
so that a representation is updated by weighting other representations according to their contextual relevance. The point of this equation here is not that attention by itself creates reference. Rather, it shows how a token such as ``Paris'' can become a context-conditioned representation. In a context about capitals, the surrounding tokens make one trajectory salient; in a context about the Louvre, another; in a context about Paris, Texas, still another. The profile that becomes active is therefore not determined by the token alone, but by the way the token is transformed through the whole available context.

This distinguishes transformer-based profile selection from a simple word-object association. The model does not first attach ``Paris'' to a fixed referent and then use that fixed referent in every continuation. Instead, the representation of ``Paris'' is repeatedly reshaped by the surrounding discourse. The relevant hidden state \(h\) at the point of prediction is already a product of this contextual transformation. It carries, in numerical form, the profile that the preceding context has made salient. This is why the model can behave as if a particular reference has been selected before the next token is produced: the selection has already been partially carried out in the formation of the context-conditioned hidden state.

The inner product then explains how such a context-conditioned structure can be converted into an output tendency. In a simplified output layer, the logit for a token \(v\) may be represented as:
\[
z_v=\langle h,e_v\rangle+b_v,
\]
where \(h\) is the current hidden state, \(e_v\) is the output vector associated with token \(v\), and \(b_v\) is a bias term. The inner product measures alignment between the current context-conditioned state and a learned output direction. In the context ``Paris is the capital of,'' the hidden state may align strongly with the output direction associated with ``France.'' In the context ``The Eiffel Tower is located in,'' it may align strongly with ``Paris.'' In the context ``Paris, Texas is,'' it should align with continuations appropriate to the Texas city rather than to the France-capital profile. The inner product does not create the referential relation. It recovers, in context, a structure that training and transformer-based contextualization have already made available.

The philosophical point is therefore not that linear algebra by itself grounds reference. It does not. The point is that once referential relations have entered language through human world-directed activity, optimization can convert those relations into distributed causal sensitivities in the model's parameter space. Transformer layers then use those sensitivities to form context-conditioned hidden states, and output alignment allows the relevant profiles to become active in token selection. In this limited but important sense, reference-bearing linguistic material can become numerically structured and contextually recoverable.

\section{Mechanistic Interpretability and Machine-Specific Reference}
Mechanistic interpretability should be understood as more than a set of suggestive visualizations. For the argument of this paper, its importance lies in intervention. If an internal feature, neuron, or activation direction merely correlates with some human-interpretable category, then the evidence remains weak: the category may be a retrospective label imposed by the investigator. But when that internal structure can be amplified, suppressed, edited, or otherwise causally manipulated, and when the model's output changes in the corresponding way, the structure is no longer merely a descriptive artifact. It is functionally active in the model's forward computation.

This provides a strong reason to say that LLMs possess reference in their own machine-specific sense. Not human reference: the model does not perceive the Golden Gate Bridge, remember Paris, understand a fact as a human knower does, or feel an emotion. But mechanistic interpretability gives evidence that parts of human referential practice become stable, recoverable, and causally active inside trained models. In this sense, entity-like features, factual neurons, and steerable activation directions are not merely evidence that the model has memorized surface strings. They are evidence of derivative referential organization: internal structures that track, preserve, and reactivate profiles inherited from human language.

The Golden Gate Bridge feature is especially important for the reference debate. Anthropic's interpretability work reports a feature in Claude 3 Sonnet associated with the Golden Gate Bridge, and shows that amplifying this feature systematically steers the model's behavior toward Golden-Gate-Bridge-related continuations \citep{Templeton2024}. The importance of this case does not lie merely in the fact that researchers assigned an interpretable label to an internal feature. It lies in the fact that the feature was both identifiable and behaviorally effective under intervention.

From the perspective of reference, the case is striking because ``Golden Gate Bridge'' is not supplied to the model as a single human concept. It reaches the model as a sequence of token-level inputs, whose exact segmentation depends on the tokenizer. The model is not given, in advance, a primitive symbol meaning the Golden Gate Bridge. Yet the relevant sequence can activate an internal structure that behaves as if the expression has been bound into a single entity-like profile. The feature is not merely about ``Golden,'' not merely about ``Gate,'' and not merely about ``Bridge.'' It is associated with the structured expression as an entity: a particular bridge, in a particular city, with a visual appearance, a name, a location, a tourist profile, and a network of descriptions.

This point directly challenges a simple word-object picture of reference. If reference were modeled only as a link between an isolated token and an object, the Golden Gate Bridge case would be difficult to understand. The relevant internal structure is activated by a multi-token expression whose components must be combined in the right way. The expression functions, at the level of the model's internal organization, as more than the sum of its token parts. It is evidence that token sequences can be transformed into stable, entity-like referential profiles.

This point should not be reduced to the surface fact that the phrase occurs frequently in the training data. A merely token-atomic picture would treat ``Golden,'' ``Gate,'' and ``Bridge'' as separate items whose effects are later added together. But the philosophical significance of this case is precisely that the ordered expression can behave as a bound unit at the level of the model's internal organization. The relevant structure is not simply the sum of a ``Golden'' profile, a ``Gate'' profile, and a ``Bridge'' profile. Through repeated exposure to contexts in which the expression is used to refer to a particular landmark, the model can stabilize a distinct entity-like profile associated with the whole expression: San Francisco, the bay, a suspension bridge, a visual appearance, tourism, maps, photographs, and other surrounding descriptions. The sequence therefore functions not merely as a string of tokens, but as a contextually bound referential profile.

This is why the case is stronger than ordinary phrase association. The issue is not simply that the model has encountered the words together many times. The issue is that the multi-token expression can activate a structure that behaves like an internal handle for an entity. If machine-specific reference exists, this is the kind of form it should take: not a primitive word-object attachment, and not a human concept, but a numerically stabilized profile produced from repeated linguistic contexts in which an expression is used as a name for something.

The generative behavior of the model provides a particularly clear illustration of the point. Once a Golden Gate Bridge profile has been activated by the input context, the model does not treat the subsequent output as an unconstrained sequence of token choices. The activated profile constrains the trajectory of generation itself. If the model begins to reproduce the name, the production of ``Golden'' strongly favors ``Gate,'' and the production of ``Golden Gate'' strongly favors ``Bridge.'' This is not because the first token already contains the whole entity as a primitive symbol. Rather, the preceding context has activated an entity-like profile whose influence persists across successive stages of generation. The output sequence therefore provides a direct illustration of computational causation: an activated profile systematically constrains what can coherently follow. In this sense, the case does not merely reveal a stored association. It reveals a causally active numerical structure that links context, activation, and output.

The causal intervention is crucial. If the feature merely correlated with occurrences of the phrase ``Golden Gate Bridge,'' the result would remain compatible with a weak surface-pattern interpretation. But when amplifying the feature makes the model produce Golden-Gate-Bridge-related outputs across contexts, the feature is shown to participate in the model's forward computation. It is not merely a label imposed by the investigator after the fact. It is a machine-internal handle on an entity-like profile.

This does not show that the model has a human concept of the Golden Gate Bridge. It has not seen the bridge, walked across it, remembered it, or attached personal experience to it. But it does show something philosophically important: a sequence of linguistic tokens can be numerically organized into a stable and causally manipulable structure corresponding to an entity-like referential profile. For the present argument, this is precisely the kind of evidence one should expect if LLM reference is derivative, language-mediated, and machine-specific rather than human-like.

Knowledge neurons provide a related but distinct kind of evidence. Dai et al. introduce knowledge neurons in pretrained Transformers by examining factual knowledge in BERT through fill-in-the-blank cloze tasks, identifying neurons that contribute to the expression of particular relational facts and showing that their activation is positively correlated with the model's expression of those facts \citep{Dai2022}. They also explore whether such neurons can be used to update or erase factual knowledge without full fine-tuning \citep{Dai2022}. For the present argument, the importance of this work is not that facts are literally stored in single neurons. That would be too simple. The importance is that factual-referential relations can become locally visible and causally manipulable within a distributed model.

This evidence differs from the Golden Gate Bridge case in an important respect. The Golden Gate Bridge feature concerns a relatively fixed multi-token proper name. The expression ``Golden Gate Bridge'' can be treated as a bound entity-like unit. Knowledge neurons, by contrast, concern relational facts that can be elicited through different linguistic formulations. A fact such as Paris--capital--France is not tied to only one surface string. It may be approached through prompts such as ``Paris is the capital of [MASK],'' ``The capital city of France is [MASK],'' or other paraphrastic variants. The relevant structure is therefore not merely a phrase-bound entity profile. It is a factual-referential relation that can remain stable across variations in expression.

This makes the reference-theoretic significance stronger. If the same relation can be accessed through different formulations, then the model's behavior cannot be explained only as sensitivity to a fixed token sequence. The model must, at least to some degree, preserve a structure that is invariant across paraphrase-like variation. From the perspective of this paper, this is precisely what a referential profile should do. It should allow different linguistic routes to converge on the same entity, relation, or fact. Knowledge neurons therefore provide evidence not simply of ``knowledge'' in a vague sense, but of locally exposed points in a distributed factual-referential profile.

The causal aspect is again crucial. If a neuron merely correlates with a factual prompt, the evidence remains weak. But if suppressing, amplifying, or editing the relevant neurons changes the model's ability to express the corresponding fact, then those neurons are functionally involved in the model's computation. They are not the whole fact, and they are not human understanding of the fact. They are local points of causal exposure within a broader distributed structure. A relation such as Paris--capital--France need not be represented as a symbolic proposition stored in one place. It may appear as a distributed pattern with intervention-sensitive sites through which the relation can be activated, weakened, or modified.

Emotion-related activation directions extend this argument beyond entity reference and factual relations. They matter not because they show that models feel emotions, but because they challenge a narrow and overly intellectualized picture of reference. Reference is often discussed as if it concerned only the neutral identification of objects or the statement of facts. But human language does not work in such a detached way. It does not merely name objects or assert propositions. It also expresses anger, irony, consolation, threat, affection, anxiety, politeness, reassurance, suspicion, and many other orientations through which utterances are used and interpreted. These orientations help determine what is salient, what is dangerous, what calls for sympathy, what counts as an accusation, and what kind of continuation is appropriate.

This point also bears on how intelligence itself is often imagined. Emotion is frequently treated as external to intelligence, as if intelligence consisted primarily in factual accuracy, logical inference, or neutral problem solving. But in human cognition, affective orientation is not simply noise added to thought. It participates in attention, salience, evaluation, memory, decision, and communication. In language, this means that affective orientation is not merely an external coloring added after reference has already been fixed. It is one of the ways in which referential profiles are selected, weighted, and pragmatically oriented.

Recent interpretability work on emotion concepts in LLMs is important in this respect. It suggests that models can contain internal representations associated with emotion concepts, and that such representations can generalize across contexts and influence behavior under activation steering \citep{Sofroniew2026}. The philosophical significance of this result is easily misread. It should not be taken as evidence that the model feels anger, fear, joy, or affection. Rather, it is evidence that the linguistic patterns through which humans express, interpret, and respond to affective orientations can be stabilized as machine-internal structures.

This expands the reference argument. A referential profile is not exhausted by an entity label or a factual relation. In ordinary discourse, the same entity or event may be referred to under different affective and pragmatic orientations. A sentence about a death may function as consolation, accusation, bureaucratic report, sarcasm, or threat. A sentence about Paris may evoke tourism, fear, nostalgia, political anger, or aesthetic admiration depending on context. These differences do not merely change the emotional tone of an already fixed reference. They help determine how the utterance is to be understood, what continuations are appropriate, and which profile is being activated.

Emotion-related activation directions therefore provide evidence for a broader form of machine-specific referential organization. Golden Gate Bridge-like features show that multi-token expressions can become bound into entity-like profiles. Knowledge neurons show that factual-referential relations can become locally exposed and causally manipulable across different formulations. Emotion-related directions show that affective-pragmatic orientations can also become steerable parts of the model's internal organization. Together, these findings suggest that interpretability research should not be read merely as a search for isolated curiosities or human-like concepts inside models. It should be read as evidence that human linguistic practice is numerically structured in LLMs across several layers: entities, facts, relations, stances, evaluations, affective framings, and discourse orientations.

This does not collapse machine reference into human experience. The model does not feel the affective orientation it tracks. But the absence of feeling does not make the internal structure irrelevant. If steering an activation direction changes the model's tone, stance, or emotional framing in systematic ways, then the direction is functionally active in the production of discourse. It is a machine-specific affective-pragmatic profile: not an emotion, not a human attitude, and not a subjective experience, but a numerically stabilized structure that tracks how emotional and pragmatic orientations are expressed in human language.

Together, these interpretability findings should be read as more than a collection of isolated curiosities. Their importance is not merely that researchers have found surprising human-interpretable features inside models. From the perspective of the vector grounding problem, they point toward a common conclusion: human linguistic practice can become numerically organized inside LLMs in ways that are stable, recoverable, and causally active. Golden Gate Bridge-like features show that multi-token expressions can be bound into entity-like profiles. Knowledge neurons show that factual-referential relations can become locally exposed and manipulable across different formulations. Emotion-related directions show that affective and pragmatic orientations can also become steerable parts of the model's internal organization.

This layered picture changes how interpretability research should be understood. It is not simply a search for hidden human concepts inside machines. Nor is it merely a catalog of correlations between internal activations and human labels. Its deeper significance is that it reveals how linguistic form can be transformed into numerical structure. Entity profiles, factual relations, and affective-pragmatic orientations do not appear in the model as human memories, perceptions, or conscious attitudes. They appear as distributed activation patterns, directions, coefficients, and intervention-sensitive sites through which trained systems select and continue discourse.

This is why these findings matter for reference. They do not prove human reference. They do not show that the model sees, remembers, understands, or feels as a human being does. But they do undermine the claim that LLMs manipulate only empty forms. If internal structures can be detected, steered, and causally linked to entity-related, fact-related, or stance-related behavior, then the relevant question is no longer whether there is a single word-object hook hidden inside the model. The question is how referential profiles inherited from human language become numerically structured and contextually reactivated. In this sense, mechanistic interpretability provides indirect but powerful evidence for machine-specific referential organization.

\section{Conclusion}
The vector grounding problem is a necessary reframing of older debates about symbol grounding. If LLMs compute over vectors rather than classical symbols, then the question of meaning and reference cannot be asked only at the level of words, symbols, or verbal outputs. It must also be asked at the level of numerical structure. A vector grounding problem requires a vector-level account of grounding.

This paper has argued that such an account must begin with a richer theory of reference. Reference should not be understood primarily as a single hook between an isolated word and an isolated object. That picture captures some important cases, especially proper names and relatively clear cases of object- or kind-reference, but it is too thin for ordinary language and especially too thin for LLMs. Even in the human case, reference is not publicly available as a private mental attachment. I may have first-person access to my own memories, images, and acts of understanding, but I do not inspect those of other speakers. I attribute reference to others through their ability to use expressions, distinguish cases, accept corrections, explain relations, reidentify topics, and maintain shared lines of discourse. Public reference therefore already has a profile-like structure: it consists in coordinated patterns of use, correction, contrast, inference, affective orientation, and continuation.

This is why the move from hooking onto to referential profiles is not merely a terminological preference. A referential profile is the form reference takes when it is considered as a communicative and context-sensitive practice rather than as a private inner image. Human beings realize such profiles through perception, memory, embodiment, affect, testimony, and social responsibility. But linguistic data contain the traces of human practices in which such profiles have been stabilized, transmitted, corrected, and reused. The question is therefore not whether LLMs possess human reference. The question is whether they can inherit parts of these publicly stabilized profiles from language and make them usable in their own computational medium.

The answer, I have argued, must be mathematical. Language-mediated reference becomes available to an LLM only by being numerically parameterized. Worldly entities and events shape human world-directed practice; human practice shapes language; language shapes training data; optimization shapes parameters; parameters shape hidden states, logits, and outputs. This mediated causal-informational chain does not establish human understanding, but it explains how world-involving linguistic practice can become a source of numerical structure inside the model. Weights, activations, attention-mediated hidden states, softmax-trained contrasts, and inner-product alignments are not secondary implementation details. They are the places where inherited linguistic relations become computationally active.

This also explains why large-scale data and mechanistic interpretability matter. Large-scale data expose the model to the many contexts in which expressions are used, contrasted, corrected, and continued. Mechanistic interpretability then provides intervention-based evidence that some of these inherited structures can become stable and causally active: entity-like features, factual-referential relations, and affective-pragmatic orientations can be detected, steered, or locally manipulated inside trained models. Such findings should not be read merely as isolated curiosities or as evidence of hidden human concepts. They are better understood as evidence that human linguistic practice can become numerically structured in LLMs across several layers.

None of this shows that mathematics creates reference from nothing. Nor does it show that LLMs perceive, remember, intend, feel, or refer as humans do. The source of grounding remains human world-directed practice, sedimented in language. But once that practice has entered linguistic form, optimization can transform its recurrent relations into stable, distributed, and causally active vector structures.

One final implication is methodological. LLMs should not be anthropomorphized, even negatively. It is misleading to ask first whether a model contains a human-like act of understanding behind its outputs, and then, when it does not, to conclude that only empty statistical mimicry remains. An LLM is a software system: it executes trained transformations over structured inputs. This does not make it a human speaker, but neither does it reduce its outputs to meaningless marks or frequency effects. The appropriate level of analysis is therefore neither hidden human mentality nor bare surface patterning, but the machine-specific organization by which inherited linguistic relations are stabilized, selected, and reactivated in numerical form. For this reason, the older Turingian framing of machinery and intelligence is more apt here than the anthropomorphic image often encouraged by ``artificial intelligence'': the issue is not artificial humanity, but machine-specific organization.

The better question is therefore not whether a human-like act of understanding is hidden behind the model's sentences, nor whether an isolated token is attached to an object by a single internal hook. The better question is how referential profiles inherited from human language become numerically stabilized and contextually recoverable in a vector system. That is the sense in which derivative, language-mediated LLM reference, if it exists, is genuinely vector grounding.

\end{document}